\documentclass{article}

 \usepackage[preprint]{neurips_2026}

\usepackage[utf8]{inputenc} 
\usepackage[T1]{fontenc}    
\usepackage{hyperref}       
\usepackage{url}            
\usepackage{booktabs}       
\usepackage{amsmath}        
\usepackage{amsfonts}       
\usepackage{nicefrac}       
\usepackage{microtype}      
\usepackage{xcolor}         
\usepackage{graphicx}

\title{Sample-Efficiency of Kolmogorov–Arnold Networks}

\author{%
  Kevin Riehl \\
  ETH Zürich, IVT \& \\
  Agentic Systems Lab (ASL) \\
  \texttt{kriehl@ethz.ch} \\
  \And
  Shaimaa K. El-Baklish \\
  ETH Zürich, IVT \\
  \texttt{selbaklish@ethz.ch} \\
  \And
  Fan Wu \\
  ETH Zürich, Agentic Systems Lab (ASL) \\
  \texttt{fanwu@ethz.ch} \\
  \And
  Anastasios Kouvelas \\
  ETH Zürich, IVT \\
  \texttt{kouvelas@ethz.ch} \\
}

\begin{document}

\maketitle

\begin{abstract}
Deep reinforcement learning has achieved substantial performance gains over classical control approaches.
Yet, a central challenge to learning in real-world applications is acquiring costly samples.
Kolmogorov-Arnold Networks are a recently proposed architecture that can learn physical relationships in control problems effectively, with significantly higher parameter efficiency and interpretability when compared to Multi-Layer-Perceptron architectures.
In this work, we systematically study sample-efficiency using computational experiments, covering the Feynman dataset and the Gymnasium RL benchmark.
The results show that similar performance can be achieved with 40\% fewer samples using the Kolmogorov-Arnold architecture, and that relative performance improvements up to 50\% occur during the training process.
The observed gains are robust to varying levels of noise in rewards.
These results highlight the potential of the Kolmogorov-Arnold architectures for more sample-efficient reinforcement learning.
\\Code: \url{https://github.com/DerKevinRiehl/neurips26\_kan\_training}
\end{abstract}

\section{Introduction}

Reinforcement learning (RL) has become a powerful paradigm for learning feedback controllers directly from system interactions~\cite{sutton2018reinforcement}.
Classical control methods, including model-based and model predictive control, can be highly effective when accurate model dynamics and tractable objectives are available, but these assumptions are often difficult to satisfy in complex, nonlinear, and noisy environments~\cite{kiumarsi2017optimal}.
Value-based RL methods such as Q-learning instead learn an action-value function that estimates the long-term return of taking an action in a given state~\cite{watkins1992qlearning}.
Deep RL extends this idea by replacing tabular value functions or policies with neural networks, enabling controllers to approximate high-dimensional and nonlinear state-action relationships~\cite{mnih2015humanlevel,schulman2017ppo}.
This flexibility has made neural RL attractive for control tasks in robotics, networking, and other cyber-physical systems where the controller must cope with sensor noise, nonlinear dynamics, delayed effects, and changing operating conditions.

For RL-based control, the dominant practical challenge is sample efficiency: learning a useful controller often requires many environment interactions~\cite{zhang2021sample}.
Unlike labeled examples in supervised learning, each control sample is a transition generated by running a policy in a simulator, digital twin, or physical system~\cite{tao2022digital}.
This makes exploration costly, since the agent must discover better actions while maintaining stable behavior under a changing data distribution~\cite{Zhangetal2021,eschmann2021reward}.
In safety-critical settings, this challenge is amplified because neural policies are difficult to inspect, verify, or translate into controller equations~\cite{Lietal2025}.
These constraints motivate function approximators that learn strong policies from fewer interactions while retaining enough structure for analysis and interpretation~\cite{ferrari2005smooth}.

Kolmogorov-Arnold Networks (KANs) are a recently proposed class of neural network architectures inspired by the Kolmogorov-Arnold representation theorem, which states that any continuous multivariate function can be expressed as a finite superposition of univariate functions and summation operators \cite{kolmogorov1956representation, arnold1963representation}.
Unlike traditional neural networks such as multilayer perceptrons (MLPs), which rely on linear transformations followed by fixed nonlinear activation functions, KANs replace scalar weights with learnable univariate functions defined along the edges of the network~\cite{liu2024kan}.
This formulation allows the model to directly learn functional relationships between variables rather than approximating them only through weighted sums of fixed activation functions.
As a result, KANs offer improved interpretability and significantly higher parameter efficiency while retaining strong approximation capabilities.
Recently, KANs have gained considerable attention, particularly in areas where understanding the underlying functional structure of the data is important \cite{somvanshi2025survey,essahraui2025kanoverview,sohail2024training}.
KANs are therefore particularly well suited for applications such as symbolic regression, scientific discovery, and physics-informed modeling, where both data efficiency and model transparency are essential \cite{somvanshi2025survey,essahraui2025kanoverview}.
These properties have also motivated the use of KANs in RL-based control, including PPO-based continuous control, robotic policy learning, load balancing, and explainable RL \cite{kich2024online,bayeh2026kanppo,chen2025kanpolicy,singh2026loadbalancing,dalto2025explainable}.

Existing evidence on the efficiency of KANs is promising.
Prior work reports that smaller KANs can match or outperform larger MLPs in several data-fitting and scientific-computing examples, suggesting favorable neural scaling behavior and parameter efficiency \cite{liu2024kan}.
In structured settings, KANs have been shown to reduce the need for expensive high-fidelity data and to perform well in low-data industrial fault detection \cite{howard2024multifidelity,villagomez2025kanae,wang2025quadkan}.
Theoretical work also gives sample-complexity guarantees for residual KANs under smoothness and noiseless-sample assumptions \cite{kratsios2025besov}.
Yet, in more systematic low-data comparisons, MLPs with individualized learnable activations can outperform KANs when only a few hundred samples are available \cite{pourkamali2025lowdata}.
These mixed findings, together with the absence of a systematic assessment of sample efficiency in RL-based control applications, motivate this work: it remains unclear when KANs improve sample efficiency over MLPs under comparable experimental budgets.

Using the Feynman dataset (a collection of physical equations) and the Gymnasium RL benchmark (a collection of RL environments)~\citep{towers2026gymnasium} we systematically assess sample efficiency of KANs as function approximators when compared to MLPs. 
Our results show that 40\% fewer samples are required to achieve similar performance, and that 50\% reward gains can be achieved during the training process using KANs.

\section{Related Works}

\textbf{Kolmogorov--Arnold networks and learning guarantees.}
KANs were introduced as an alternative to MLPs in which learnable univariate functions are placed on edges rather than fixed nonlinearities on nodes \cite{liu2024kan}.
This architecture has motivated rapid follow-up work because it connects neural approximation to interpretable function decompositions and can use fewer parameters in some scientific-computing and symbolic-regression settings \cite{somvanshi2025survey,essahraui2025kanoverview}.
Theoretical work has begun to formalize these intuitions by proving approximation and sample-complexity guarantees for residual KANs on smooth function classes (including Besov spaces) \cite{kratsios2025besov}.
These guarantees are important, but they are typically stated under assumptions that differ from RL control, where targets are noisy, bootstrapped, and induced by a changing policy.

\textbf{Low-data and fixed-sample empirical comparisons.}
Empirical evidence for KAN sample efficiency remains mixed.
The original KAN experiments show favorable scaling and strong performance with compact models, but they do not isolate environment-sample efficiency under fixed compute and fixed interaction budgets \cite{liu2024kan}.
Follow-up works directly study low-data regimes and find that MLPs with individualized learnable activations can outperform KANs with substantially fewer parameters, especially when only a few hundred samples are available \cite{pourkamali2025lowdata}.
Further studies show that KANs and MLPs trade advantages across regular, irregular, discontinuous, and noisy functions, with KANs converging quickly in some cases but failing to dominate across stabilized test loss \cite{zeng2024irregular}.
These comparisons suggest KANs should not be treated as universally more sample-efficient than MLPs.

\textbf{Structured and expensive-data settings.}
More positive results appear when the problem structure matches the inductive bias of KAN edge functions.
Some works propose multifidelity KANs that combine abundant low-fidelity data with limited high-fidelity data, reducing the amount of expensive high-fidelity data needed in scientific machine learning \cite{howard2024multifidelity}.
Furthermore, studies evaluate KAN autoencoders for fault detection across varying training-set sizes and find that several KAN variants match or surpass an MLP-based orthogonal autoencoder with substantially less data \cite{villagomez2025kanae}.
These results are highly relevant for control domains with costly simulation or measurement, but they are supervised or unsupervised learning studies rather than online RL studies.

\textbf{KANs in RL and control.}
Several recent works integrate KANs into RL pipelines.
Prior works replace MLP actor and critic components in PPO with KANs and report comparable performance with fewer parameters on DeepMind Control Proprio robotics benchmarks \cite{kich2024online}.
Similarly, studies propose KAN-PPO and MultKAN-PPO for continuous control, finding robust performance across several configurations while noting sensitivity to dropout design \cite{bayeh2026kanppo}.
Other studies focus on KAN-based PPO agents for CartPole and radio-access-network control, showing that KAN agents can match MLP performance with far fewer parameters and can be converted into sparse symbolic policies \cite{dalto2025explainable}.
Moreover, works propose a KAN actor in PPO for load balancing and emphasize the ability to extract interpretable controller equations \cite{singh2026loadbalancing}.
Together, these works support the feasibility of KAN-based RL, but they primarily emphasize parameter efficiency and interpretability rather than controlled sample-efficiency comparisons.

\textbf{KANs for robotic policies and world models.}
KANs have also been used in policy-learning systems outside the standard actor-critic replacement setting.
KAN Policy can be used for diffusion-based visuomotor policy learning, and works doing so report smoother trajectories, shorter execution times, and improved success rates relative to diffusion-policy baselines \cite{chen2025kanpolicy}.
\textit{QuadKAN} for vision-guided quadruped locomotion reports faster convergence, improved returns, lower action jitter, and better robustness in structured gait-control settings \cite{wang2025quadkan}.
\textit{KAN-Dreamer} \cite{shi2025kandreamer} replaces selected components with KAN or FastKAN \cite{li2024kolmogorovarnold} modules, finding that FastKAN is promising for reward and continue prediction but that direct replacement of visual encoders or actor-critic modules can harm sample efficiency.
This component-level evidence reinforces the need to distinguish where KANs are useful in a RL stack rather than assuming that every MLP can be replaced beneficially.

\section{Experiments}

\subsection{Feynman Sample-Efficiency Benchmark}
The first experiment tests the core approximation question in a controlled supervised setting, before moving to full RL control.
If KANs are more sample-efficient because their edge functions provide a useful inductive bias, this advantage should already be visible when learning smooth physical laws from few samples.
For this purpose, we follow Liu et al. and use the Feynman benchmark, which contains closed-form physics equations derived from the Feynman Lectures and related symbolic-regression benchmarks \cite{liu2024kan,udrescu2020aifeynman,udrescu2020aifeynman2}.
Each benchmark task is a regression problem, where each Feynman Lecture equation is independently sampled to generate training sets (varying size) and test sets (fixed size).
The primary evaluation metric is normalized root mean squared error (NRMSE).

\subsection{RL Gymnasium Benchmark}
The second experiment studies the sample efficiency of KANs and MLPs in the context of DeepRL for control.
For this purpose, we use the RL benchmark "\textit{Gymnasium}"~\citep{towers2026gymnasium}, which contains multiple different physics-based environments to study RL.
We chose the subset of "Classic Control" as state variables are used for learning, rather than image or video streams, as KANs are known to be particularly suitable to learn physical relationships.
Furthermore, we chose the proximal policy optimization (PPO) learning strategy, as it is a standard, robust policy-gradient method for continuous and discrete control problems~\citep{schulman2017proximal}.
The benchmark includes the following environments:
"Acrobot-v1", "CartPole-v1", "MountainCar-v0", "MountainCarContinuous-v0", and "Pendulum-v1".
Each benchmark task provides a comparable learning curve with best rewards achievable (convergence limits) enabling a comparison.

\subsection{RL Gymnasium Benchmark With Noised Rewards}
The third experiment studies the robustness of the sample efficiency gains of KANs over MLPs in the context of DeepRL in control applications, for varying reward (Gaussian) noise levels.

\subsection{Comparison and Sample-Efficiency Assessment}
Different model complexities and training parameters were explored via grid-search to identify optimal performance.
KAN and MLP models were matched by model complexity (number of parameters) to ensure a fair comparison.
Number of training samples to achieve a certain performance were used to assess the enhanced sample-efficiency of KANs over MLPs.
Specifically, we compare ReLU and Sigmoid for MLP, and B-Splines and radial basis functions (RBF) for KAN networks, to cover a multitude of different function space shapes.

\section{Results}

\begin{figure}[!h]
    \centering
    \includegraphics[width=0.8\linewidth,trim={0 0.1cm 0 0},clip]{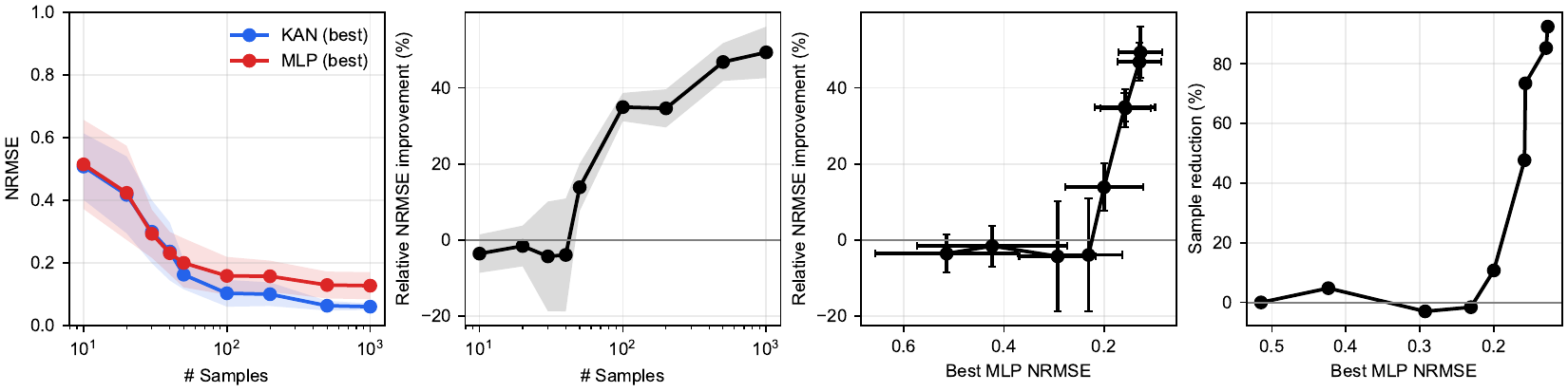}
    \caption{Experiment 1: Sample Efficiency in Feynman Benchmark.}
    \label{fig:result_exp1}
\end{figure}

%
%
The Feynman experiment shows a clear low-sample crossover in favor of KANs after 50 training samples (Fig.~\ref{fig:result_exp1}).
The best KAN family reduces mean NRMSE from 0.200 to 0.163 (at 50 samples) and from 0.127 to 0.060 (at 1000 samples), corresponding to a relative improvement from 13.9\% (at 50 samples) to 49.3\% (at 1000 samples).
Equivalently, KANs reach several MLP error levels with substantially fewer training samples once the target error is below the early-data regime.

\begin{figure}[!h]
    \centering
    \includegraphics[width=0.8\linewidth,trim={0 0.8cm 0 0},clip]{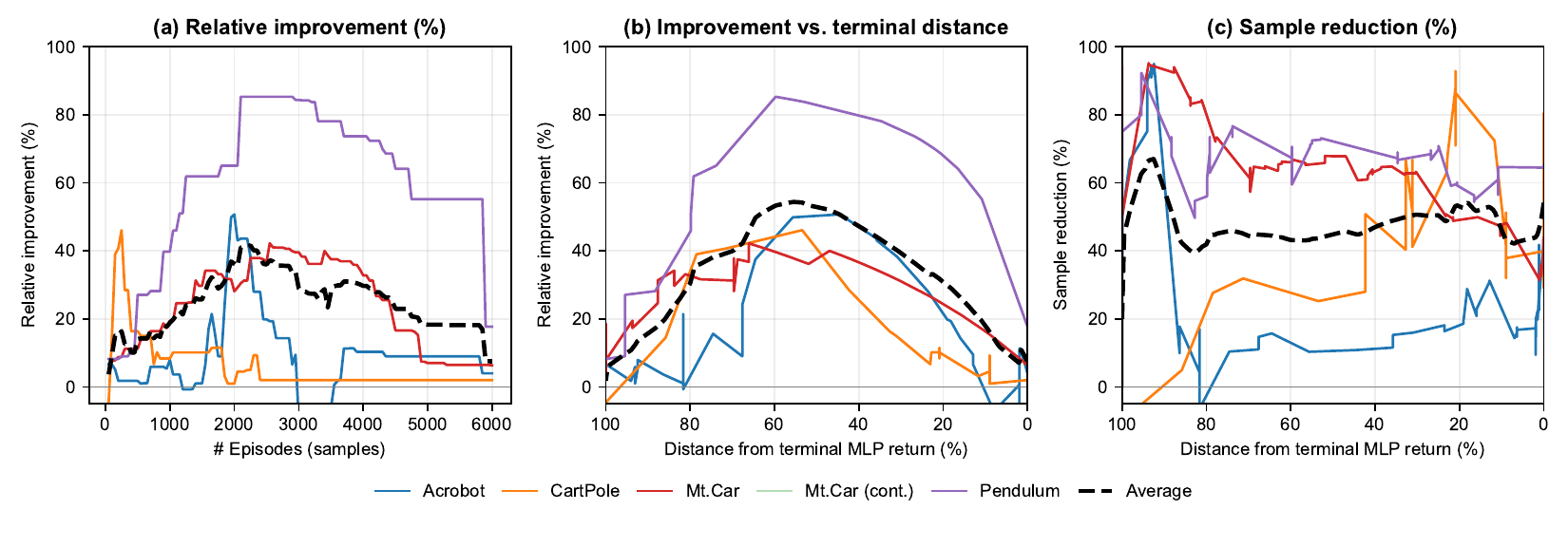}
    \caption{Experiment 2: Sample Efficiency in Gymnasium Benchmark.}
    \label{fig:result_exp2}
\end{figure}

\begin{figure}[!h]
    \centering
    \includegraphics[width=0.8\linewidth]{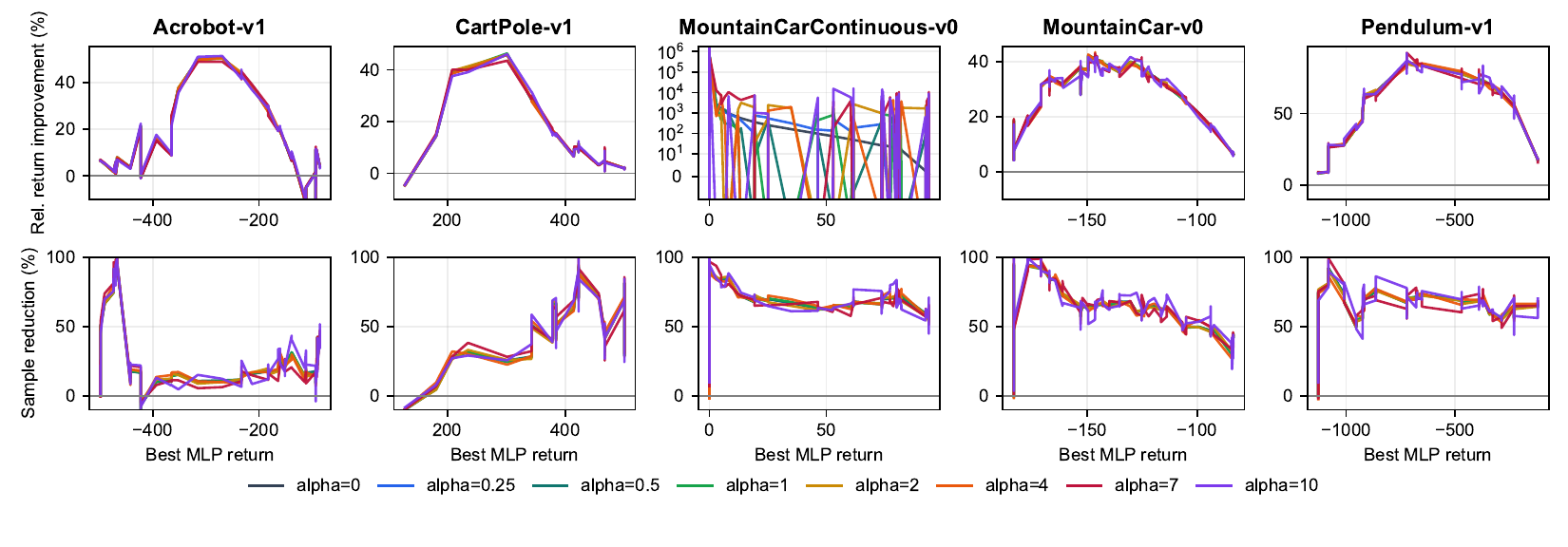}
    \caption{Experiment 3: Sample Efficiency in Noised Gymnasium Benchmark.}
    \label{fig:result_exp3}
\end{figure}

The Gymnasium experiment (Fig.~\ref{fig:result_exp2}) reproduces the same pattern in control applications, although the size of the effect depends strongly on the environment.
Across Acrobot, CartPole, MountainCar, and Pendulum, the advantage of KANs peaks at about 41.6\% (2250 episodes) and then narrows to about 7.5\% (final checkpoint) on average.
The sample-reduction curves show that that similar performances can be achieved with about 40\% fewer samples using KANs.
After many training episodes, both architectures approach similar return levels, demonstrating the faster learning ability of KANs over MLPs.
The noisy-reward benchmark indicates that these gains are not an artifact of clean reward feedback.
Adding Gaussian reward noise increases variance, but the KAN and MLP comparison keeps the same qualitative shape across the tested noise scales (Fig.~\ref{fig:result_exp3}).

\section{Discussion}

Overall, the experiments highlight that KAN architectures improve sample efficiency when the task contains smooth low-dimensional structure.
The strongest gains occur in the early and middle part of training, which is the regime that matters most when simulator calls, measurements, or hardware trials are expensive.
The effect is not uniform across tasks, and the final returns are often close once both model architectures receive many samples.
The results suggest that KANs can be useful for low-data control problems.
A practical route for deployment is to train with KANs during the scarce-sample phase and then distill the learned policy into a cheaper MLP when inference cost is the bottleneck.

\newpage


\bibliographystyle{unsrt}
\bibliography{references}

\clearpage
\appendix

\section{Experiment Setup Details}

\subsection{Feynman Benchmark Details}

\subsubsection{Dataset Construction}

The Feynman benchmark consists of analytic equations with named input variables and prescribed sampling ranges \cite{udrescu2020aifeynman,udrescu2020aifeynman2}.
Following Liu et al., we evaluate on 27 implemented `Feynman\_no\_units' equations for which finite train and test samples can be generated \cite{liu2024kan}.
This includes univariate equations, which are reported separately from multivariate equations in the aggregate analysis.
We use the unit-free subset, because univariate equations provide a sanity check for spline-based edge learning, while multivariate equations test whether KANs retain any sample-efficiency advantage when the model must learn interactions between variables.
For each equation $f_j:\mathbb{R}^{d_j}\rightarrow\mathbb{R}$, we generate supervised examples by sampling $x\sim\mathcal{U}(\mathcal{D}_j)$ and evaluating $y=f_j(x)$.
The domain $\mathcal{D}_j$ is the Cartesian product of the variable intervals specified by the benchmark.
For each equation, one fixed train pool and one fixed test set are sampled, and model initializations are varied across random seeds.
If an equation is undefined at a sampled point or returns a non-finite value, we reject the point and resample until the desired split size is reached.
This rejection step is applied before normalization and is identical for all models.

\subsubsection{Sample-Efficiency Protocol}

For each equation, we fix $N_{\mathrm{test}}=1000$.
We vary the training size over
\begin{equation}
N_{\mathrm{train}}\in\{10,20,30,40,50,100,200,500,1000\}.
\end{equation}.
For each $N_{\mathrm{train}}$, equation, and model family, we repeat the experiment over $20$ random seeds.
Each seed determines the training draw, the test draw, and the model initialization.
Inputs are standardized using the mean and standard deviation of the training set.
Targets may be standardized during optimization, but all reported metrics are computed after transforming predictions back to the original target scale.
No information from the test set is used for model selection or normalization.

\subsubsection{Model Comparisons}

The main comparison includes KAN and MLP regressors trained with the same optimizer, loss, maximum number of update steps, batch size, and random seeds.
For KANs, we use spline-based edge functions as in the original KAN implementation, with spline order and grid size fixed across equations unless otherwise stated.
For MLPs, we compare ReLU and sigmoid activations, to avoid giving KANs an artificial advantage over nonsmooth baselines on smooth physics equations.
We report trainable parameter counts and wall-clock training time for every trained model.
The primary setting uses a fixed architecture recipe for each model family.
A secondary parameter-matched setting chooses MLP widths so that the number of trainable parameters is as close as possible to the KAN parameter count.
When exact matching is impossible, we report the parameter ratio alongside the result.
This protocol avoids treating parameter count as interchangeable across architectures while still checking whether apparent sample-efficiency gains are explained by model size.

\subsubsection{KAN Implementations}

To validate KAN baselines, we use the public All-KAN implementation repository as the implementation source for several KAN variants.
The implementation collection is organized around the original KAN model and the KAN 2.0 scientific-discovery framework \cite{liu2024kan,liu2024kan2}.
Our experiment wrapper currently exposes EfficientKAN for B-spline KANs, FastKAN for Gaussian RBF edge functions, and BSRBF-KAN for hybrid B-spline/RBF edge functions \cite{li2024kolmogorovarnold,ta2024bsrbf}.
The All-KAN repository also includes recent variants such as FC-KAN, PRKAN, and AF-KAN, which provide implementation references for future ablations \cite{ta2024fc,ta2025prkan,ta2025afkan}.
All implementation choices are fixed across sample budgets within an experiment, and each reported result specifies the backend used to instantiate the KAN model.

\subsubsection{Metrics}

Let $\{(x_i,y_i)\}_{i=1}^{N_{\mathrm{test}}}$ be the test set for one equation and let $\hat{y}_i$ be the model prediction.
We define the test-set mean and standard deviation as
\begin{equation}
\bar{y}=\frac{1}{N_{\mathrm{test}}}\sum_{i=1}^{N_{\mathrm{test}}}y_i,
\qquad
s_y=\sqrt{\frac{1}{N_{\mathrm{test}}}\sum_{i=1}^{N_{\mathrm{test}}}(y_i-\bar{y})^2}.
\end{equation}
The root mean squared error is
\begin{equation}
\mathrm{RMSE}=\sqrt{\frac{1}{N_{\mathrm{test}}}\sum_{i=1}^{N_{\mathrm{test}}}(\hat{y}_i-y_i)^2}.
\end{equation}
The normalized root mean squared error is
\begin{equation}
\mathrm{NRMSE}=\frac{\mathrm{RMSE}}{s_y+\epsilon},
\end{equation}
where $\epsilon=10^{-12}$ prevents numerical instability for nearly constant targets.
The mean absolute error is
\begin{equation}
\mathrm{MAE}=\frac{1}{N_{\mathrm{test}}}\sum_{i=1}^{N_{\mathrm{test}}}|\hat{y}_i-y_i|.
\end{equation}
The relative $\ell_2$ error is
\begin{equation}
\mathrm{Rel}\text{-}\ell_2=\frac{\sqrt{\sum_{i=1}^{N_{\mathrm{test}}}(\hat{y}_i-y_i)^2}}{\sqrt{\sum_{i=1}^{N_{\mathrm{test}}}y_i^2}+\epsilon}.
\end{equation}
The coefficient of determination is
\begin{equation}
R^2=1-\frac{\sum_{i=1}^{N_{\mathrm{test}}}(\hat{y}_i-y_i)^2}{\sum_{i=1}^{N_{\mathrm{test}}}(y_i-\bar{y})^2+\epsilon}.
\end{equation}
RMSE and MAE preserve the physical scale of each equation.
NRMSE and relative $\ell_2$ error allow comparisons across equations with different output magnitudes.
$R^2$ reports explained variance and is easier to interpret when communicating whether a model has learned the dominant structure of the function.

\subsubsection{Aggregate Reporting}

For each equation and training size, we report the mean and standard error over seeds.
Across equations, the plotted summaries average errors within each seed and report mean and standard deviation across seeds, since a small number of high-scale or ill-conditioned equations can dominate aggregate errors.
We report these aggregate statistics for all equations, for univariate equations only, and for multivariate equations only.
The full sample-efficiency curve plots NRMSE against $N_{\mathrm{train}}$ on a logarithmic x-axis.
As scalar summaries, we report relative improvement and sample reduction against the best MLP curve.
The area under the error-sample curve is computed by trapezoidal integration of NRMSE over $\log N_{\mathrm{train}}$, where lower values indicate better sample efficiency.
The threshold $N_\tau$ is the smallest training-set size for which the median NRMSE falls below a chosen target $\tau$.
If no method reaches $\tau$, we report $N_\tau>1000$.

\subsection{RL Gymnasium Benchmark Details}

\subsubsection{Benchmark and Agents}

We use the Gymnasium Classic Control environments CartPole-v1, Acrobot-v1, MountainCar-v0, MountainCarContinuous-v0, and Pendulum-v1.
The observations are low-dimensional state vectors, which matches the intended use of KAN edge functions for smooth physical relationships.
The runner supports both discrete and continuous action spaces, using categorical policies for discrete actions and diagonal Gaussian policies with tanh squashing for continuous actions.
For continuous actions, the trainable log standard deviation is initialized to $0$.

\subsubsection{Model Sweep}

The main comparison uses full actor-critic agents, so the actor and critic are both KANs for KAN runs and both MLPs for MLP runs.
The KAN sweep includes B-spline KANs and Gaussian RBF KANs with hidden dimensions $[2]$, $[4]$, $[8]$, $[4,4]$, $[8,8]$, and $[16,16]$.
The MLP sweep includes ReLU and sigmoid MLPs with hidden dimensions $[16]$, $[32]$, $[64]$, $[32,32]$, and $[64,64]$.
The KAN settings use grid size $5$, spline order $3$, grid range $[-1.5,1.5]$, and SiLU as the base activation.
The runner records actor, critic, and total trainable parameter counts for every checkpoint.

\subsubsection{PPO Training and Evaluation}

Each training job collects complete episodes, computes generalized advantage estimates with $\gamma=0.99$ and $\lambda=0.95$, and applies clipped PPO updates with clip parameter $0.20$.
The optimizer is Adam with learning rate $3\times10^{-4}$, no weight decay, entropy coefficient $0.01$, value coefficient $0.50$, four PPO epochs per update, and gradient clipping at $0.50$.
Updates are performed after batches of five episodes in the full protocol.
Training seeds are paired across model families, and evaluation uses deterministic actions with five evaluation episodes per checkpoint.
Discrete policies are evaluated with the action of maximum logit, while continuous policies are evaluated with the mean action.
The result file used for the paper contains ten paired random seeds and learning curves up to 6000 training episodes.
During training, episodes, cumulative environment steps, recent training returns, deterministic evaluation returns, episode lengths, PPO losses, runtime, and parameter counts are logged.

\subsection{RL Gymnasium Noise Benchmark Details}

\subsubsection{Reward Perturbation}

The noisy benchmark keeps the same Gymnasium environments, PPO update, model sweep, and deterministic clean evaluation protocol as the clean Gymnasium benchmark.
During training only, each clean reward $r_t$ is replaced by $r_t^{\mathrm{noise}}=r_t+\alpha\epsilon_t$, where $\epsilon_t\sim\mathcal{N}(0,1)$.
The main sweep uses $\alpha\in\{0,0.25,0.5,1,2,4,7,10\}$ in raw reward units.
Because noise only enters the rewards used for advantage estimation, the experiment tests robustness to corrupted training feedback rather than robustness to stochastic evaluation.

\clearpage
\section{Detailed Results}

\begin{figure}[!h]
    \centering
    \includegraphics[width=0.8\linewidth]{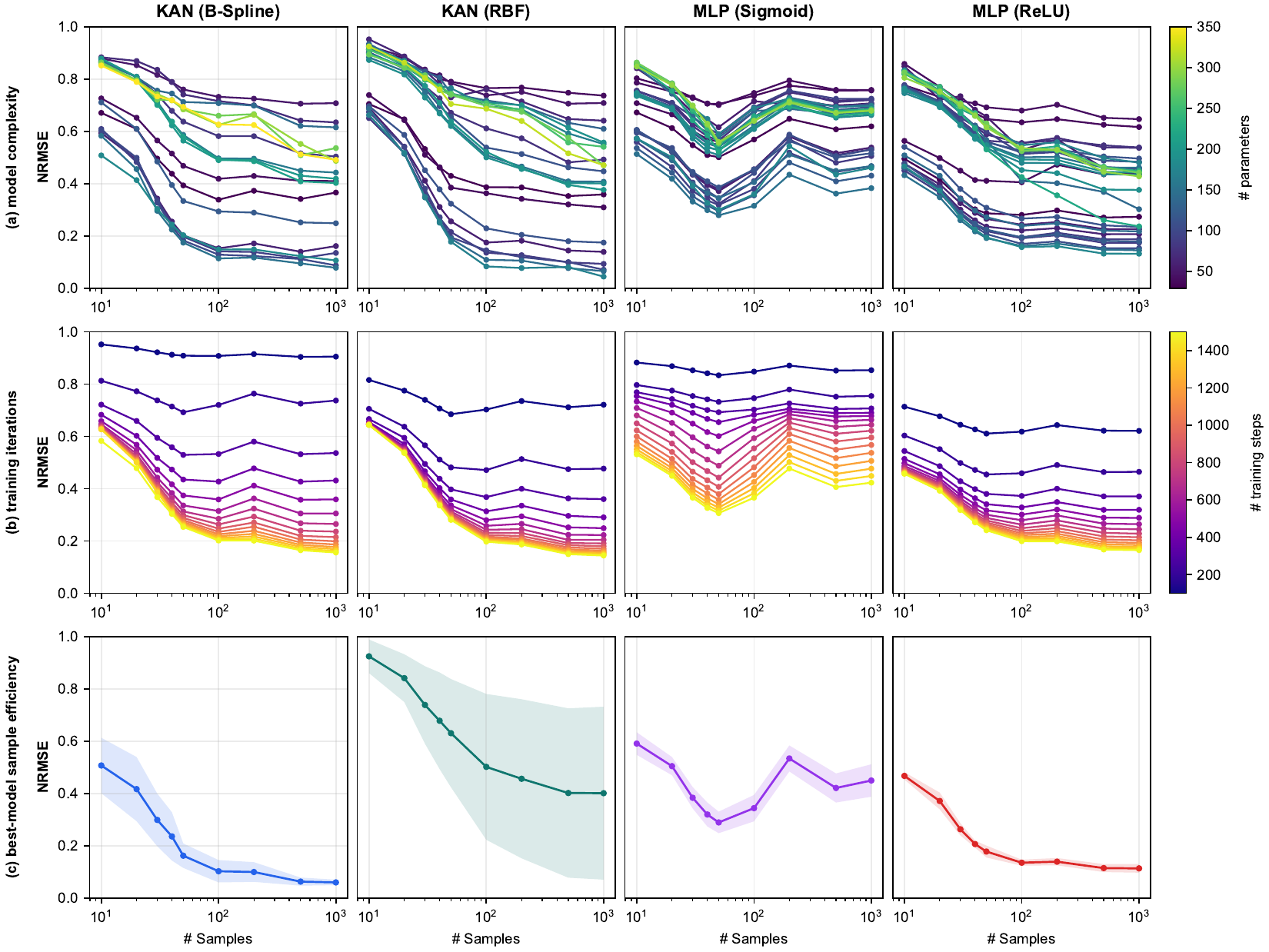}
    \caption{Experiment 1: Feynman Benchmark Experiment Design.}
    \label{fig:experiment_1_results}
\end{figure}

\begin{figure}[!h]
    \centering
    \rotatebox{270}{\includegraphics[width=1.2\linewidth]{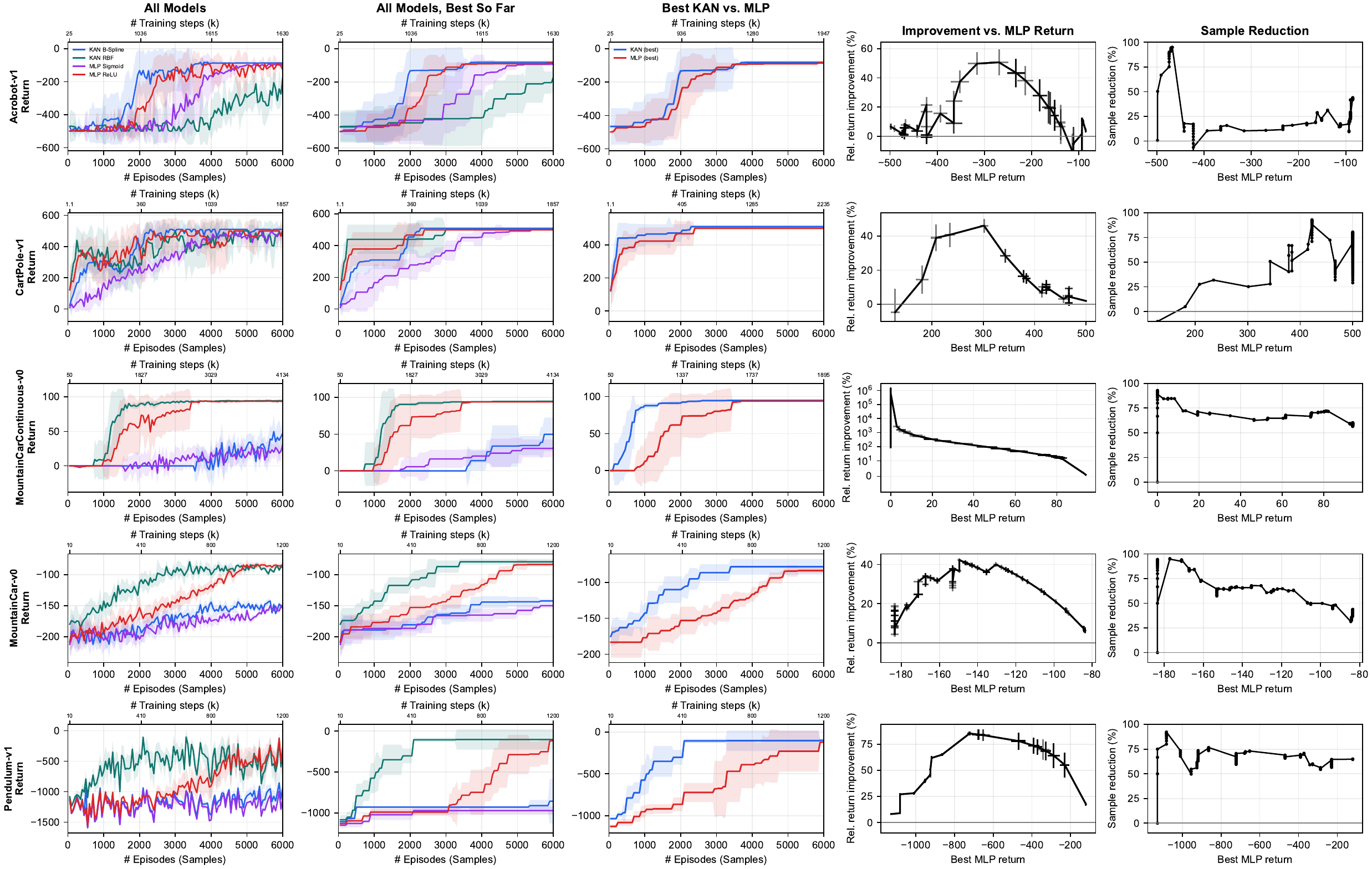}}
    \caption{Experiment 2: DeepRL Gymnasium Experiment Design.}
    \label{fig:experiment_2_results}
\end{figure}

\end{document}